\documentclass{article}

\usepackage[numbers]{natbib}
\usepackage[final]{neurips_2023}

\usepackage[dvipsnames]{xcolor}         
\definecolor{linkColor}{rgb}{0.18,0.39,0.90}
\usepackage[utf8]{inputenc} 
\usepackage[T1]{fontenc}    
\usepackage[colorlinks=true,linkcolor=linkColor,citecolor=linkColor,filecolor=linkColor,urlcolor=linkColor]{hyperref}       
\usepackage{url}            
\usepackage{booktabs}       
\usepackage{amsfonts}       
\usepackage{nicefrac}       
\usepackage{microtype}      
\usepackage{xcolor}         
\usepackage{natbib}
\usepackage{listings}
\usepackage[utf8]{inputenc} 
\usepackage[T1]{fontenc}    
\usepackage{booktabs}       
\usepackage{amsfonts}       
\usepackage{nicefrac}       
\usepackage{microtype}      
\usepackage{xcolor}         
\usepackage{pifont}
\usepackage{threeparttable}
\usepackage{multirow}
\usepackage{amsfonts}
\usepackage{bbm}
\usepackage{bm}
\usepackage{bbold}
\usepackage{color}
\usepackage{graphicx}
\usepackage{amsmath}
\usepackage{subfigure}
\usepackage{enumitem}
\usepackage[noend]{algorithm2e}
\usepackage{wrapfig}
\input{Definitions}
\usepackage[most]{tcolorbox}

\newcommand{\bblue}[1]{{\textbf{\color{blue}{#1}}}}
\newcommand{\bred}[1]{{\textbf{\color{red}{#1}}}}
\newcommand{\bblack}[1]{{\textbf{\color{black}{#1}}}}

\newcommand\ourf{\makebox{\textsc{MLM-Filter-GPT4}}}
\newcommand\ours{\makebox{\textsc{MLM-Filter-GPT4V}}}

\allowdisplaybreaks[4]

\begin{document}

\title{Finetuned Multimodal Language Models Are High-Quality Image-Text Data Filters}
%

\author{Weizhi Wang$^1$ \quad Khalil Mrini$^2$ \quad Linjie Yang$^2$ \quad Sateesh Kumar$^2$\\
\textbf{Yu Tian$^2$ \quad Xifeng Yan$^1$ \quad Heng Wang$^2$}  \\
$^1$University of California, Santa Barbara~~~$^2$Bytedance, US \\\\
\url{https://mlm-filter.github.io}
}

\maketitle
\begin{abstract}

We propose a novel framework for filtering image-text data by leveraging fine-tuned Multimodal Language Models (MLMs). Our approach outperforms predominant filtering methods (\eg CLIPScore) via integrating the recent advances in MLMs. We design four distinct yet complementary metrics to holistically measure the quality of image-text data. A new pipeline is established to construct high-quality instruction data for fine-tuning MLMs as data filters. Comparing with CLIPScore, our MLM filters produce more precise and comprehensive scores that directly improve the quality of filtered data and boost the performance of pre-trained models. We achieve significant improvements over CLIPScore on popular foundation models (\ie CLIP and BLIP2) and various downstream tasks. Our MLM filter can generalize to different models and tasks, and be used as a drop-in replacement for CLIPScore. An additional ablation study is provided to verify our design choices for the MLM filter.

\end{abstract}

\section{Introduction}



Large-scale image-text datasets~\citep{cc3m,cc12m,laion400m,laion5b,coyo700m} have been the major driving force for the recent breakthrough in Vision-Language Models (VLMs) and Text-to-Image generation models. The ever-growing size of such datasets allows researchers to scale the models to unprecedented capacities with billions or even trillions of parameters. These humongous foundation models lead to significant improvements in many down-stream tasks, such as image classification, text-to-image retrieval, image captioning, visual question answering, image generation and editing, \etc One great example is the OpenAI CLIP~\citep{clip} model, which is trained with 400M web-crawled image-text pairs. The CLIP model demonstrates impressive zero-shot learning capability across a wide range of different tasks.





The quality of image-text data plays a decisive role in the final performance of foundation models. But web-crawled image-text data are often very noisy, \eg the corresponding text data is low quality or does not match the content of the image. How to build high-quality image-text datasets is a challenging research problem that attracts lots of interests recently. \cite{xu2023demystifying} try to re-create the data curation process from CLIP. \cite{nguyen2022quality} advocate that data quality is more important than quantity for model robustness. The \textsc{DataComp} challenge~\citep{datacomp} is introduced to systematically evaluate different data-filtering techniques.

\begin{figure}[t] 
\centering 
\includegraphics[width=0.8\textwidth]{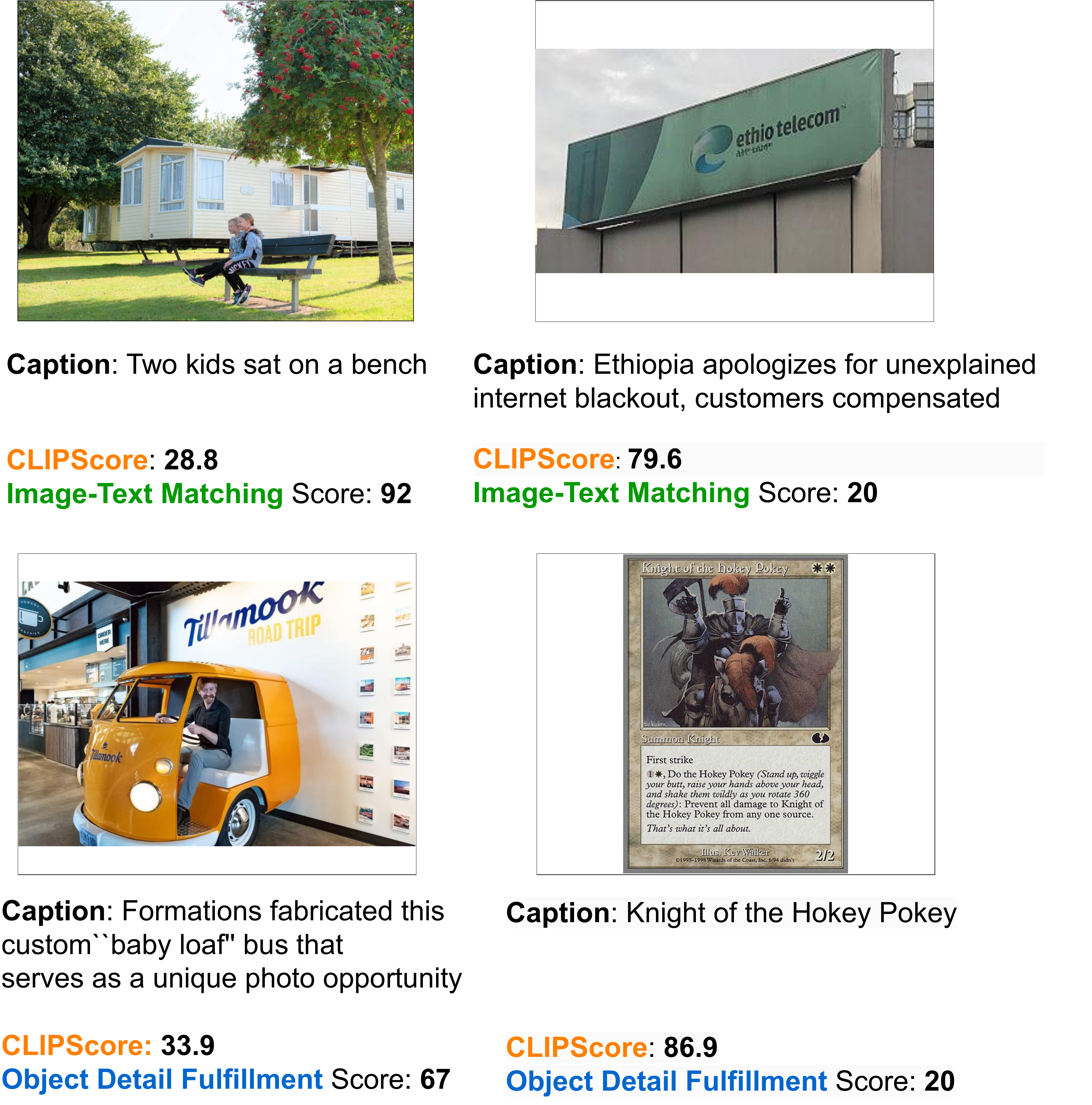} 
\caption{CLIPScore fails in differentiating the fine-grained object-level image-text alignment, while the image-text matching score generated by MLM Filter significantly captures such alignment.
}
\label{fig:showcase}
\end{figure}

Each successful foundation model have their own secret recipes for data filtering. Before the invention of CLIP, most techniques are hand-designed or rule-based. For example, CC3M and CC12M design a series of heuristics for image-based, text-based and image\&text-based filtering. Model-based filtering becomes popular since the introduction of CLIPScore~\citep{hessel2021clipscore}, which leverages the CLIP model to compute the cosine similarity between image and text to measure their alignment. CLIPScore has become the predominant method for filtering image-text data. 
However, recent research~\citep{tong2023mass,tong2024eyes} finds that visual features from CLIP are blind to subtle differences in the image, \eg object number, shape and position. Because the contrastive loss is applied to the whole image, CLIPScore is less sensitive to capture the fine-grained object-level alignment information, shown in Figure~\ref{fig:showcase}.
Additionally, the text encoder of CLIP can only process up to 77 tokens. The information loss from the text encoder can limit CLIPScore to process data with long captions. This limitation can be serious for Text-to-Image generation models~\citep{dalle3} that rely on long and highly-descriptive captions.

Compared with the contrastively trained CLIP model, Multimodal Language Models (MLMs) have demonstrated promising capability in predicting the quality of generated images or text and aligning well with human preferences. 
More specifically, the image-text matching scores generated by GPT-4Vision~\citep{gpt4v} are more consistent with human experts compared with CLIPScore in recent MLM-based evaluation~\citep{msgpt4v,g4veval}.  
This motivates us to integrate recent advances in MLMs for high-quality data filtering:
\begin{quote}\itshape
    ``Can we adapt strong MLMs to generate scores for assessing image-text data quality and outperform CLIPScore for image-text data filtering?''
\end{quote}



Though GPT-4V is better at measuring image-text alignment, directly applying GPT-4V-scale MLMs in filtering billions of image-text data is computationally too costly. A good filtering method should be both effective and efficient due to the sheer amount of data we need to process.
There are smaller MLMs (\eg LLaVA~\citep{llava}, MiniGPT-4~\citep{zhu2023minigpt}, etc), which are more efficient but fail to generate scores at a granularity that can reflect the subtle changes in the image-text data, since they are mainly instruction-tuned on task completion data. In this paper, we propose to combine the best of both worlds, leveraging proprietary LLMs or MLMs to construct high-quality instruction tuning data for effectiveness, and fine-tuning more accessible open-source MLMs to inject the knowledge from the high-quality data for efficiency.

We summarize our major contributions as follows:
\begin{itemize}[leftmargin=*]
    \item We propose the MLM filter which incorporates the recent progress from MLMs for image-text data filtering and can be used as a drop-in replacement to the popular CLIPScore.  
    \item We design four diverse metrics to measure the image-text data quality from different perspectives, and a new pipeline to construct high-quality instruction data to harvest the information from proprietary models.
    \item Foundation models trained with our MLM filtered data demonstrate significant improvements, \eg  1.7\% better on 38 downstream tasks from \textsc{DataComp} comparing with CLIPScore.
    
\end{itemize}




\section{Related Work}
\textbf{Data Filters.} Initial work, such as ImageNet~\citep{deng2009imagenet}, relies on manual data filtering to select high-quality images and captions. More recent work~\citep{clip,jia2021scaling} pushes the size of image-text dataset to the order of hundreds of millions, and thus employs fixed rules and heuristics for filtering. LAION~\citep{laion400m} introduce the CLIPScore metric computed by the pre-trained CLIP model in filtering high-quality image-text pairs. CLIPScore filtering then becomes a widespread method of constructing large-scale web-crawled datasets~\citep{coyo700m,laion5b,datacomp}. Based on that, \textsc{DataComp} \citep{datacomp} is the first work to propose a benchmark for evaluating data filtering methods. \cite{yu2023devil} introduce a set of tools to improve data filtering including CLIP-FLIP, distribution matching, de-duplication and clustering. Similarly, \cite{maini2023t} propose text masking to improve filtering. On the other hand, \cite{fang2023data} use high quality image-text pairs to train a new CLIP filtering network instead of using OpenAI's original CLIPScore. These papers all build upon CLIP filtering and introduce various techniques to improve it. In contrast, we investigate an alternate approach to CLIP-based Filtering, which employs fine-tuned Multimodal Language Models for large-scale image-text data filtering. Additionally, various works~\citep{chen2023alpagasus,wei2023instructiongpt} deploys proprietary LLMs like GPT-4 to score and filter text-only and visual instruction data.

\textbf{Multimodal Language Models.} Recent Multimodal Language Models~\citep{flamingo,kosmos-1,valm,blip2,zhu2023minigpt,llava} concatenate vision encoders with the latest LLMs via cross-model adapters to enable LLMs~\citep{alpaca,vicuna,llama2} to take visual inputs. The most typical vision encoders deployed in MLMs are still the vision transformer models in CLIP pre-trained models~\citep{clip} for extracting visual features of input images. Moreover, various adapter architectures are proposed to connect the feature space of different modalities, including Q-former proposed by BLIP-2~\citep{blip2}, a simple MLP layer used in LLaVA~\citep{llava}, and Visual Experts of CogVLM~\citep{cogvlm}.

\textbf{Multimodal Instruction Tuning.} Instruction tuning~\citep{mishra2021cross,flan,instructgpt} is a fine-tuning paradigm that enables LLMs to perform unseen tasks. This zero-shot performance is enabled by training LLMs using natural language instructions to explain the goal of the task. Instruction tuning is much more computationally efficient than full-set fine-tuning, and can enable LLMs to achieve zero-shot performance scores that are competitive with fully supervised models. LLaVA~\citep{llava} introduces multimodal instruction tuning via fine-tuning MLMs on a set of visual instructions. MLMs that use instruction tuning~\citep{instructblip,llava2} achieve SOTA performance on various vision-language tasks, such as visual question answering and visual reasoning.

\section{Fine-Tuned Multimodal Language Models as Data Filters}

\subsection{Overview}
\begin{figure}[ht] 
\centering 
\includegraphics[width=0.8\textwidth]{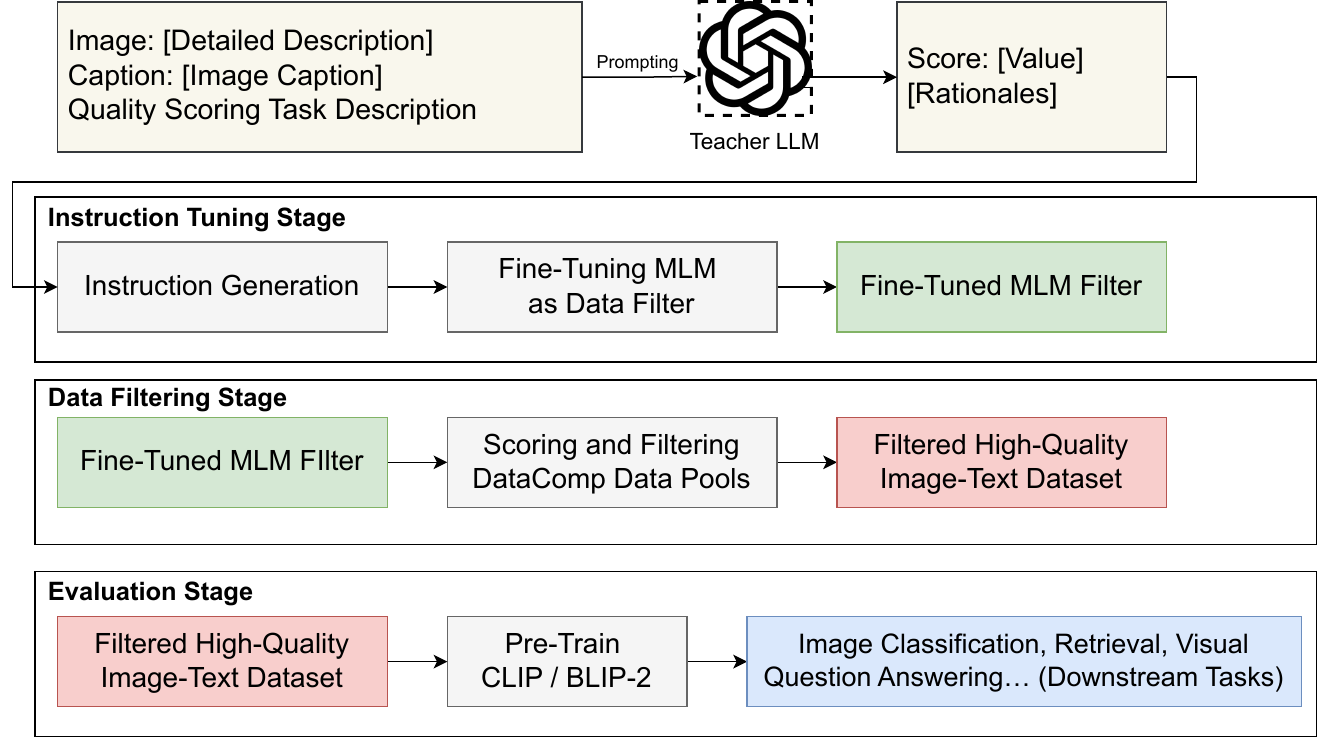}
\caption{Illustration of the pipeline of fine-tuning MLM Filter and employing it for data filtering.}
\label{fig:pipeline}
\end{figure}

We propose to adopt fine-tuned Multimodal Language Model as effective data filters to select high-quality image-text data to promote the VLM pre-training, which involves three stages: 1) constructing multimodal instruction tuning data on proposed quality scoring tasks to fine-tune MLM to realize accurate quality assessment; 2) adopt the fine-tuned MLM Filter to generate quality scores for each data point in the data pool and then select the high-quality data; 3) pre-train VLMs using the filtered dataset and evaluate the pre-trained VLMs on downstream tasks to demonstrate the effectiveness of the proposed filtering method. The detailed pipeline for the three stages is shown in Figure~\ref{fig:pipeline}.



\subsection{Constructing Multimodal Instruction Tuning Data for Scoring Tasks}
\label{sec:data_construct}

In order to work as an effective data filter, the MLM must generate quality scores for every single image-text pair for data selection and filtering. To enable MLMs like LLaVA to reason accurately on the quality score, we propose to fine-tune such MLMs on a set of scoring tasks to enhance their scoring capability. The multimodal instruction tuning data needed for scoring tasks are hard and expensive to collect via human labeling, and thus we leverage proprietary models GPT-4 or GPT-4V to construct such multimodal instruction data for scoring tasks.

\textbf{Defining Metrics for Image-Text Quality Assessment.}
Conventional data filters like CLIPScore focus on the overall holistic matching of image and text via computing the cosine similarity between hidden features of image and text. However, such implicit scoring is poor in discriminating hard or ambiguous samples, leading to the false negative score predictions shown in Figure~\ref{fig:showcase}. We propose to leverage strong Multimodal Language Models to predict the quality scores towards image-text pairs. Beyond the overall image-text alignment assessment, the fine-tuned MLM filters can evaluate the quality of image-text pairs from multiple perspectives. We propose four quality evaluation metrics to comprehensively evaluate the data quality: 
\begin{itemize}[leftmargin=*]
    \item Image-Text Matching (ITM): the ITM metric focuses on evaluating whether the image caption accurately represents the main features and objects of the image and captures its primary theme. The fine-tuned MLM data filter can explicitly generate the ITM score on a scale of 100.
    \item Object Detail Fulfillment (ODF): the ODF metric focuses on evaluating whether the image caption provides detailed descriptions of objects that align with the image. Specifically, ODF assesses if the caption sufficiently describes the properties of the objects in the image, \eg number, color, size, position, shape, etc. Compared with the ITM metric, the ODF metric focuses more on the fine-grained alignment between the detailed object properties in the image and the ones described in the corresponding caption.
    \item Caption Text Quality (CTQ): the CTQ metric focuses on evaluating the text quality of image caption based on the grammatical correctness, diversity of vocabulary (\eg the range and uniqueness of words), fluency (\eg smoothness and natural flow of sentences), readability, length, and structure. Previous data-centric research~\citep{yu2023devil} finds that web-crawled data is poor in its text quality, as it contains various bad text patterns, such as repeated words or textual noise. Thus, we propose to fine-tune MLMs to assess the text quality of image captions for data filtering.
    \item Semantic Understanding (SU): the SU metric focuses on determining if the image caption provides additional semantic information that is not readily apparent just from the image itself. Such auxiliary semantic information can be 1) the professions of persons in the image; 2) the locations, addresses, festivals, country names, city names; 3) the names or entities of buildings, people, bird species, animal breeds, car models, engines in the image; 4) the social relationships between the people in the image, \ie lovers, parent, or child. We suggest that adopting SU metric for data filtering can select image-text pairs with auxiliary semantics, which can further enhance the commonsense reasoning capability of pre-trained VLMs.
\end{itemize}



\textbf{Prompting the Teacher Models.} We select two state-of-the-art teacher models, GPT-4 and GPT-4V, to construct the multimodal instruction data for quality scoring tasks. Constructing multimodal instruction data with GPT-4V is much easier as GPT-4V can directly take visual inputs. As GPT-4 is a text-only LLM, we transform the image into a detailed text description to prompt a text-only GPT-4. The prompt for such dense captioning process is \textit{Please generate a dense caption in 4-6 sentences for describing the image in detail as much as you can}. These comprehensive image descriptions are generated using a SOTA image captioning models, such as LLaVA or ShareGPT4V~\citep{chen2023sharegpt4v}. With the prompt to the teacher model and the generated output, the visual instruction data can be simply formatted as \textit{User: \{Prompt\} Assistant: \{Output\}}.

\begin{figure}[t]
\centering
\subfigure[]
{
	\begin{minipage}{0.45\textwidth}
	\centering          
	\includegraphics[width=1\textwidth]{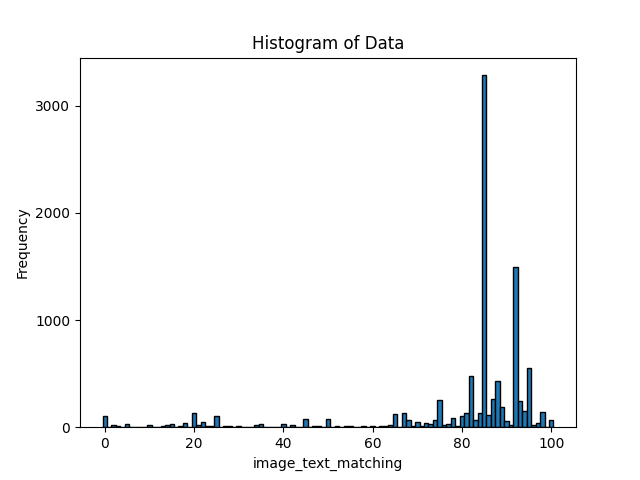}
        \label{fig:g4v_distribution_original}
        \vspace{-10pt}
	\end{minipage}
}
\subfigure[]
{
	\begin{minipage}{0.45\textwidth}
	\centering     
	\includegraphics[width=1\textwidth]{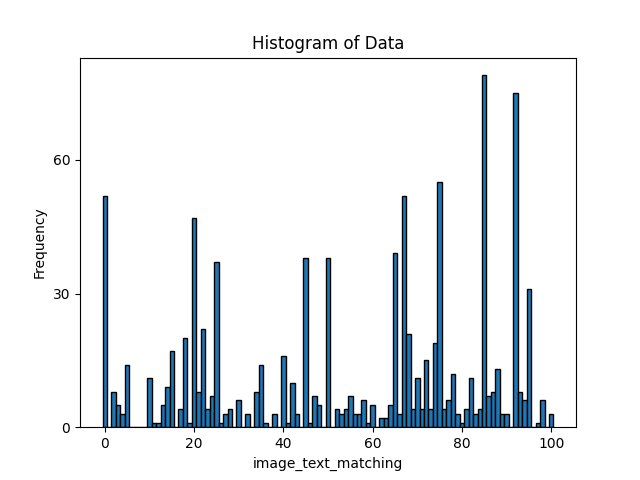}
        \vspace{-10pt}
        \label{fig:g4v_distribution_sampled}
	\end{minipage}
}
\caption{(a) image text matching score distribution of initial 10k instructions using GPT-4V on CC12M; (b) image text matching score distribution of final 1k instructions uniformly sampled from 10 buckets.}
\label{fig:ablation} 
\end{figure}

\textbf{Prompting Strategies.} As the scoring tasks involve a reasoning process to predict final accurate quality metrics for an image-text pair, we consider two prompting strategies to ensure the reasoning accuracy of the fine-tuned multimodal language model: Chain-of-Thought (CoT) Reasoning~\citep{cot}, and Rationalization Reasoning~\citep{rationalization}. The major difference between the two prompting strategies are the generation order of the score and the generated reasoning steps. The exemplar prompts for two prompting strategies are presented in Appendix~\ref{appendix:prompt_strategy} Table~\ref{tab:prompt_strategy}. Between these two prompting strategies, we select the rationalization reasoning as we find it to be the most efficient and accurate. Computational efficiency is a concern as the scoring MLM should be able to score billions of image-text pairs. If the MLM is fine-tuned to output the score value first, the model's text generation process can be stopped early in the inference stage as only the score value is needed for filtering. Secondly, the experimental results of LLaVA demonstrate that the instruction tuning with rationalization reasoning leads to better performance on the ScienceQA benchmark~\citep{saikh2022scienceqa} than CoT reasoning. Four final prompts for different scoring metrics are presented in Appendix~\ref{appendix:prompts}.


\textbf{Selecting Image-Text Pairs for Data Collection.} 
The multimodal instruction data used for fine-tuning should contain image-text pairs of varying quality. 
Thus, data diversity is essential to enhance the fine-tuned MLM filter, enabling it to effectively score image-text data across all quality levels. We select two different image-text dataset as the data pool for constructing instruction tuning data: the Conceptual Captions 12M (CC12m)~\citep{cc12m}, and the \textsc{DataComp} Medium 128M Dataset~\citep{datacomp}. 
To enhance the diversity of the instruction set, we perform clustering and uniform-sampling on the sentence embeddings of each captioning text.
The sentence embedding model we use is the pre-trained MPNet~\citep{song2020mpnet} encoder model, which is contrastively pre-trained on a mixture of retrieval and natural language inference datasets. We directly use the pre-trained MPNet provided by \texttt{Sentence Transformers}~\citep{sbert} to generate the sentence embedding towards each image caption. We set the number of clusters as $10k$ and $20k$ for CC12M and Datacomp-Medium, respectively. The image-text pairs for constructing instruction tuning data are uniformly sampled from each cluster, in which only one data point closest to the cluster centroid is selected. 

\begin{table*}[t]
\centering
\small
\scalebox{0.83}{
\begin{tabular}{@{}l l llc cccc}
\hline    
\toprule
\multirow{2}{*}{\textbf{Captioner}} & \multirow{2}{*}{\textbf{Data Resource}} & \multirow{2}{*}{\begin{tabular}[c]{@{}l@{}}\textbf{\#Sampling} \\ \textbf{Buckets}\end{tabular}} & \multirow{2}{*}{\begin{tabular}[c]{@{}l@{}}\textbf{Teacher} \\ \textbf{Model}\end{tabular}} & \multirow{2}{*}{\textbf{ImageNet-1k}} & \multirow{2}{*}{\begin{tabular}[l]{@{}c@{}}\textbf{ImageNet} \\ \textbf{dist. shifts}\end{tabular}} & \multirow{2}{*}{\textbf{VTAB}} & \multirow{2}{*}{\textbf{Retrieval}} & \multirow{2}{*}{\begin{tabular}[l]{@{}c@{}}\textbf{Average over} \\ \textbf{38 datasets}\end{tabular}} \\
    & & & & & & & \\ 
\midrule
\textbf{LLaVA} & CC12M & 10 & GPT-4 & \underline{29.0} & 24.5 & 35.0 & \underline{29.3} & \underline{34.2} \\
\textbf{ShareGPT4V} & CC12M & 10 & GPT-4 & 28.4 & \underline{24.9} & \underline{35.3} & 28.2 & 33.7 \\
\midrule
N/A & \textbf{DataComp} & 10 & GPT-4V & 29.6 & 24.8 & \underline{34.2} & 26.7 & 33.2 \\
N/A & \textbf{CC12M} & 10 & GPT-4V & \underline{30.5} & \underline{25.3} & 33.4 & \underline{28.0} & \underline{33.7}   \\
\midrule
ShareGPT4V & CC12M & \textbf{10} & GPT-4 & \underline{28.4} & \underline{24.9} & \underline{35.3} & 28.2 & \underline{33.7} \\
ShareGPT4V & CC12M & \textbf{100} & GPT-4 & 27.5 & 23.0 & 34.6 & \underline{28.8} & 33.2 \\
\midrule
LLaVA & CC12M & 10 & \textbf{GPT-4} & 29.0 & 24.5 & \underline{35.0} & \underline{29.3} & \underline{34.2}  \\
N/A & CC12M & 10 & \textbf{GPT-4V} &\underline{30.5} & \underline{25.3} & 33.4 & 28.0 & 33.7  \\
\bottomrule
\hline
\end{tabular}
}
\caption{Ablations on different design choices for constructing multimodal instruction data for quality scoring tasks.}
\label{tab:ablation}
\end{table*}

\textbf{Sampling Final Instructions for Scoring Tasks.}
As we find that the initial $10k$ instruction data generated by teacher models are not uniformly distributed on the score scale of $100$ in Figure~\ref{fig:g4v_distribution_original}, we need to sample the initial instruction data into a balanced instruction set to avoid learning bias. Considering that the ideal size of multi-task instruction tuning dataset is $50k$ instructions~\citep{vicuna,llama2}, we decide to sample $1k$ instructions from $10k$ initial generated instruction data for each scoring tasks, which ensure the generalization capability of instruction-tuned MLM. Thus, there are $4k$ instruction data of quality scoring tasks to be included in the total $50k$ instruction dataset, such that there is 1k instruction data for each proposed quality metric. We experiment with two sampling methods to ensure that the instruction data distribution is balanced on the scoring scale of $100$: 1) grouping all data into $10$ buckets and uniformly sampling $100$ instructions from each bucket; 2) grouping all data into $100$ buckets and uniformly sampling $10$ instructions from each bucket. The score distribution of sampled 10k instruction in Figure~\ref{fig:g4v_distribution_sampled} are more diverse and uniform than the original score distribution in Figure~\ref{fig:g4v_distribution_original}. The code for sampling the final $4k$ instruction is presented in Appendix~\ref{appendix:sampling}.



\textbf{Mixture with instruction data of multi-tasks.}
The multimodal instruction tuning process should involve a diverse set of tasks~\citep{instructblip,llava2} to enhance the zero-shot reasoning capability of fine-tuned MLMs. In addition to 4k multimodal instruction data of the proposed data quality scoring tasks, we sample another 46k multimodal instructions from LLaVA-665k instruction datasets. We allocate a larger portion of our data mixture to reasoning tasks, such as complex reasoning~\citep{llava} and GQA~\citep{hudson2019gqa} as we regard that enhancing reasoning capabilities will improve the scoring capability of our fine-tuned MLM. 
The detailed statistics on the size of each dataset sampled for data mixture are presented in Appendix~\ref{appendix:data_mixture} Table~\ref{tab:data_mixture}.


\subsection{Instruction-Tuning on Multimodal Language Models}
We adopt LLaVA-1.5 based on Vicuna-13B LLM~\citep{vicuna,llava2} as the Multimodal Language Model architecture for instruction tuning on the mixed instructions of data quality scoring tasks and other multimodal tasks. The training process of LLaVA-1.5 involves pre-training on image-text pairs and instruction tuning on multimodal instructions. We directly take the pre-trained checkpoint and only reimplement the instruction tuning stage with our mixed instruction set.

\subsection{Creating Optimal MLM Data Filters}
We propose various different design choices for constructing instruction data for data quality scoring tasks in Section~\ref{sec:data_construct}. These design choices may make a significant difference in the effectiveness of instruction tuning. To create the optimal fine-tuned MLM data filter, we conduct comprehensive ablation studies to investigate the effects of different design choices on the filtering performance. Four major design choices for constructing the instruction data for scoring tasks are investigated: 1) we experiment with two captioning models to transform image into text-base detailed description for prompting GPT-4, including LLaVA and ShareGPT4V~\citep{chen2023sharegpt4v}; 2) we experiment with two different image-text datasets for constructing visual instructions, including CC12M and DataComp Medium 128M; 3) we experiment with two different numbers of grouping buckets, 10 and 100, for sampling the final 4k instructions; 4) we experiment with different teacher models to get multimodal instructions, including GPT-4 and GPT-4 Vision. Additionally, we use the DataComp benchmark to evaluate the effectiveness of different data filtering hyperparameters. 

\noindent\textbf{DataComp Benchmark.} The DataComp benchmark \citep{datacomp} has been introduced to systematically compare the performance of different data filtering methods. In this benchmark, the training code and computational budget is fixed across all competing methods to facilitate direct comparison between methods. The DataComp provides a fixed original image-text data pool for different filtering methods to ensure a fair comparison. The performance is measured by training a CLIP model on the filtered dataset and then testing the zero-shot capabilities of this CLIP model on a suite of 38 classification and retrieval tasks. We select the Medium scale training setting to train ViT-B/32 CLIP models on datasets resulting from various MLM data filter configurations.
 

\begin{table*}[t]
\centering
\small
\scalebox{0.85}{
\begin{tabular}{@{}l l lcc c cc}
\hline    
\toprule
\multirow{2}{*}{\textbf{Filter}} & \multirow{2}{*}{\textbf{Metrics}} & \multirow{2}{*}{\begin{tabular}[c]{@{}l@{}}\textbf{Teacher} \\ \textbf{Model}\end{tabular}} & \multirow{2}{*}{\textbf{ImageNet-1k}} & \multirow{2}{*}{\begin{tabular}[l]{@{}c@{}}\textbf{ImageNet} \\ \textbf{dist. shifts}\end{tabular}} & \multirow{2}{*}{\textbf{VTAB}} & \multirow{2}{*}{\textbf{Retrieval}} & \multirow{2}{*}{\begin{tabular}[l]{@{}c@{}}\textbf{Average over} \\ \textbf{38 datasets}\end{tabular}} \\
    & & & & & & & \\ 
\midrule
No Filtering & - & - & 17.6 & 15.2 & 25.9 & 21.9 & 25.8 \\
Basic Filtering & Rules & - & 22.6 & 19.3 & 28.4 & 25.1 & 28.5\\
LAION Filtering & CLIPScore+Rules & - &23.0 & 19.8 & 30.7 & 23.3 & 29.2 \\
CLIPScore & CLIPScore & - & 27.3 & 23.0 & 33.8 & 25.1 & 32.8 \\
\midrule
\textsc{MLM-Filter}  & Image-Text Matching & GPT-4 & 28.6 & 23.7 & 34.4 & \textbf{30.0} & 33.4\\ 
\textsc{MLM-Filter}  & Object Detail Fulfillment &  GPT-4 & 29.0 & 24.5 & 35.0 & 29.3 & \underline{34.2} \\
\textsc{MLM-Filter}  & Caption Text Quality &  GPT-4 & 25.2 & 20.9 & 32.1 & 26.4 & 30.9  \\
\textsc{MLM-Filter}  & Semantic Understanding &  GPT-4 & 20.3 & 16.1 & 28.4 & 20.2 & 27.0 \\
\midrule
\textsc{MLM-Filter}  & Image-Text Matching & GPT-4V & 29.4 & 24.4 & \textbf{36.1} & \underline{29.7} & \underline{34.2}  \\ 
\textsc{MLM-Filter}  & Object Detail Fulfillment &  GPT-4V & \textbf{30.5} & \underline{25.3} & 33.4 & 28.0 & 33.7 \\
\textsc{MLM-Filter}  & Caption Text Quality &  GPT-4V & 24.3 & 20.4 & 32.3 & 24.5 & 30.9 \\
\textsc{MLM-Filter}  & Semantic Understanding &  GPT-4V & 16.2 & 13.9 & 23.3 & 18.7 & 24.0 \\
\midrule
\textsc{MLM-Filter}  & ITM AND ODF & GPT-4V & \underline{30.3} & \textbf{25.6} & \underline{36.0} & 29.0 & \textbf{34.5} \\
\textsc{MLM-Filter}  & ITM OR ODF & GPT-4V & 28.9 & 24.5 & 35.2 & 29.0 & 33.9 \\
\bottomrule
\hline
\end{tabular}
}
\caption{Zero-shot performance of CLIP models pre-trained using baseline filtering methods and proposed \textsc{MLM-Filter} on \textit{Medium} scale pools of the DataComp benchmark. AND represents the combination of ITM and ODF metrics using AND operation.}
\label{table:medium}
\end{table*} 

\paragraph{Ablation Results.} To investigate the effects of each design choice, we keep the selection of the other three design choices the same and only change one design choice for each experiment group. As we propose four different metrics to assess data quality, we only adopt the metric of \textit{Object Detail Fulfillment} as the filtering metric to select a high-quality subset from the 128M medium scale data pool. The ablation results for all four design choices are presented in Table~\ref{tab:ablation}.

The first two lines in Table~\ref{tab:ablation} demonstrate that adopting LLaVA as the captioning model to transform images into detailed descriptions for instruction data construction leads to better filtering performance. Next, adopting CC12M to sample image-text pairs for data construction outperforms the design choice of using DataComp-Medium dataset. We suppose it is because the image quality of CC12M is significantly better than that of DataComp, enabling the instruction tuning process more knowledge intensive. Thirdly, grouping the initial instructions into 10 buckets for sampling illustrates priority over using 100 buckets. In terms of the selection of teacher models, the MLM filters learned from different teacher models exhibit distinct strengths across different tasks. The MLM filter learned from GPT-4 performs better in VTAB~\citep{vtab} classification and retrieval datasets, while the MLM filter learned from GPT-4V obtains higher scores in ImageNet~\citep{deng2009imagenet} related datasets. Finally, we decide to fix the other three choices as LLaVA captioner, CC12M data resources, and 10 sampling buckets. We report the two versions of MLM-based filters with different teacher models GPT4 and GPT-4V for future experiments, denoted as \textsc{MLM-Filter-GPT4} and \textsc{MLM-Filter-GPT4V} respectively.



\section{Experiments}
In this section, we evaluate the effectiveness of adopting fine-tuned MLMs as high-quality image-text data filters. We compare the performance of vision-language models pre-trained on datasets filtered using a baseline filter with their performance using our MLM filter. We select two different VLM architectures for comprehensive evaluation: CLIP pre-training and BLIP-2 pre-training. Additionally, we conduct human evaluation to compute the correlation between the scoring generated by our proposed MLM filter model and the baseline CLIP model.

\subsection{CLIP Pre-Training on DataComp Medium and Large Scales}
\begin{table*}[htb]
\centering
\small
\scalebox{0.85}{
\begin{tabular}{@{}l l lcc c cc}
\hline    
\toprule
\multirow{2}{*}{\textbf{Filter}} & \multirow{2}{*}{\textbf{Metrics}} & \multirow{2}{*}{\begin{tabular}[c]{@{}l@{}}\textbf{Teacher} \\ \textbf{Model}\end{tabular}} & \multirow{2}{*}{\textbf{ImageNet-1k}} & \multirow{2}{*}{\begin{tabular}[l]{@{}c@{}}\textbf{ImageNet} \\ \textbf{dist. shifts}\end{tabular}} & \multirow{2}{*}{\textbf{VTAB}} & \multirow{2}{*}{\textbf{Retrieval}} & \multirow{2}{*}{\begin{tabular}[l]{@{}c@{}}\textbf{Average over} \\ \textbf{38 datasets}\end{tabular}} \\
    & & & & & & & \\ 
\midrule
No Filtering & - & - & 45.9 & 37.8 & 42.6 & 41.9 & 43.7 \\
Basic Filtering & Rules & - & 51.6 & 42.3 & 44.6 & 48.0 & 45.8 \\
LAION Filtering & CLIPScore+Rules & - &  55.3 & 45.3 & 51.0 & 49.5 & 50.1 \\
CLIPScore & CLIPScore & - & 57.8 & 47.4 & 53.8 & 46.6 & 52.9 \\
\midrule
\textsc{MLM-Filter} & Object Detail Fulfillment &  GPT-4 &  58.9 & 48.9 & 57.4 & 52.5 & 54.2 \\
\bottomrule
\hline
\end{tabular}
}
\caption{Zero-shot performance of CLIP models pre-trained using baseline filtering methods and proposed \textsc{MLM-Filter} on \textit{Large} scale pools of the DataComp benchmark.}
\label{table:large}
\end{table*} 


\textbf{Evaluation Setup.} We select the DataComp benchmark to evaluate the effectiveness of adopting fine-tuned MLM as data filter. The evaluation process involves the data filtering stage and evaluation stage, which are shown in Figure~\ref{fig:pipeline}. During the data filtering stage, we adopt the MLM-Filter to generate quality scores on all 128M medium-scale data and 1.28B large-scale data. After that, an integer filtering threshold is calculated based on the closest value that retains 30\% of the overall data pool, 38.4M for Medium and 384M for Large. Such threshold is set up to select all the image-text pairs, of which the quality score is larger or equal to the threshold. 
We report the results using each defined metric to filter data separately and we consider two MLM filters learning from different teacher models. Additionally, we also report the results of experiments with a combination of two metrics for data filtering. 
Finally, we select a high-quality subset from the medium or large scale image-text data pools based on different proposed quality metrics. During the evaluation stage, we adopt the selected high-quality data subset to pre-train a CLIP model and compare the performance of our CLIP model with CLIP models pre-trained on datasets filtered by other methods. 

\textbf{Baselines.} We compare the proposed MLM filter with other baseline filtering methods from DataComp, including applying no filtering, basic filtering, LAION filtering and CLIPScore filtering. The basic filtering method adopts three rule-based filters, filtering English only, filtering by caption length, and filtering by image size. The LAION filtering adopts both the CLIPScore filtering using ViT-B/32 CLIP model and the English filtering. The CLIPScore filtering utilizes a larger ViT-L/14 CLIP model for score generation and data filtering.

\textbf{Training Details.} 
We strictly follow the training setup provided by DataComp. The computational budget and hyperparameters are fixed for pre-training CLIP using different filters. The CLIP model architecture is determined by the data scale, in which the ViT-B/32 model is pre-trained on the medium scale setting and ViT-B/16 model is on the large scale setting. We use $32$ Nvidia A100 GPUs to train our models.


\textbf{Results on DataComp Medium and Large Scale.} The DataComp results between the proposed MLM filter and other baselines are presented in Table~\ref{table:medium} and Table~\ref{table:large} for Medium and Large scale respectively. On the medium-scale DataComp benchmark, the proposed MLM Filter significantly outperforms the CLIPScore baseline on different task subgroups, achieving notable improvements of +3.2 accuracy on ImageNet-1k, +2.6 average accuracy on 6 ImageNet shifted datasets, +2.3 average accuracy on 13 VTAB datasets, and +4.9 average scores on 3 retrieval datasets. Moreover, the proposed \textsc{MLM Filter} surpasses CLIPScore baseline by +1.7 and +1.3 improvements on the average scores over 38 datasets on DataComp Medium and Large Scale benchmarks, which demonstrates the proposed MLM Filter can work as more effective filtering method than CLIPScore filter. Additionally, we can draw the following auxiliary conclusions from the results:

\textbf{The MLM Filter learned from GPT-4V performs better on ImageNet related datasets than the MLM Filter learned from GPT-4.} The \ours{} achieves the best performance on both ImageNet-1k and 6 ImageNet Shifted datasets. Both filtering metrics of Image Text Matching and Object Detail Fulfillment generated by \ours{} outperforms the best ImageNet-1k accuracy of \ourf{}, achieving a notable improvement of +1.1 accuracy. 

\textbf{The optimal filtering metric varies for fine-tuned MLM Filter learned from different teacher models.} For the proposed \textsc{MLM Filter} learned from different teacher models, the optimal filtering metric under single metric filtering setting is different. The Image-Text Matching is the optimal filtering metric for \ours{}, while the Object Detail Fulfillment metric helps the \ourf{} most. The other two metrics of Caption Text Quality and Semantic Understanding cannot work as effective filtering quality metrics in DataComp benchmark, leading to worse performance than CLIPScore baseline. We regard that it is because the most of DataComp evaluation datasets are image classification datasets, which did not aligh with the filtering directions and objectives of CTQ and SU metrics.

\textbf{Image-Text Matching is the best filtering metric for retrieval tasks.} Our proposed MLM Filter achieves the SOTA performance on the three image-to-text and text-to-image datasets under DataComp Medium setting. The two types of MLM Filters achieves 30.0 and 29.7 average performance on three retrieval tasks using the ITM filtering metric, surpassing the CLIPScore baseline by 4.9 average scores. We also observe in results of both MLM Filter variants that the image-text matching metric leads to better performance on retrieval tasks compared with other three filtering metrics.

\textbf{Combing different quality metrics effectively filters and identifies image-text pairs of better quality. } The AND operation to combine ITM and ODF quality metrics means that the ITM and ODF score of selected datapoints should exceed the filtering thresholds of both metrics, while the OR operation to combine two metrics means that the selected datapoints should either exceed the threshold for ITM metric or that for ODF metric. The combination of ITM and ODF metrics using AND operation outperforms all the baseline filtering methods and other variants of MLM Filters, achieving the best average performance of 34.5 over 38 datasets.

\begin{table*}[!t]
\centering
\small
\scalebox{0.95}{
\begin{tabular}{p{29mm}p{8mm}p{8mm}p{8mm}p{8mm}p{8mm}l l lll}
\hline
\toprule
\textbf{Filter} &\textbf{Metrics} & \textbf{SVHN}& \textbf{MNIST}& \textbf{Avg.} \\ 
\midrule
\ourf{} & ITM & 8.2 & 10.3 & 9.2  \\
\ourf{} & ODF & 14.6 & 19.3 & 16.9 \\
\midrule
\ours{}  & ITM & 15.4 & 8.3 & 11.8 \\
\ours{} & ODF &  9.0  & 6.8 & 7.9 \\
\midrule
\ours{} & AND & 12.9 & 11.6 & 12.3 \\
\bottomrule
\hline
\end{tabular}
}
\caption{
Zero-shot performance of pre-trained CLIP on SVHN and MNIST digit classification datasets. Avg. represents the average performance on two digit datasets. AND represents the combination of ITM and ODF metrics using AND operation.
}
\label{tab:digit}
\end{table*}

\textbf{The worse performance on digit classification tasks prevents \textsc{MLM-Filter-GPT4V} from remarkably outperforming \ourf{}.} Even if \textsc{MLM-Filter-GPT4V} outperforms \ourf{} on 23 ImageNet, VTAB and retrieval datasets, it only achieves the same average performance over 38 datasets as \textsc{MLM-Filter-GPT4}. It is because the performance of \textsc{MLM-Filter-GPT4V} on the two digit classification datasets significantly lags behind \ourf{} by 5.1 average score, shown in Table~\ref{tab:digit}, which leads to 0.27 average score behind on 38 datasets. The combination of two quality metrics promotes the digit classification performance of \ours{}, but does not resolve it.

\subsection{BLIP2 Pre-Training}
To demonstrate the effectiveness of our proposed MLM Filter across various VLM model architectures, we pre-train BLIP-2 VLM on the filtered dataset and evaluate the zero-shot performance of such BLIP-2 model on VQA datasets to compare the effectiveness of filtering methods on high-level vision-language tasks.

\textbf{Training setup. }  We directly use the filtered dataset from DataComp Large 1.28B data pool using CLIPScore filtering and our proposed MLM Filtering. The batch size and number of pre-training steps are kept as the same as original implementation~\citep{blip2} for both the CLIPScore filtered dataset and MLM filtered dataset, in which both BLIP-2 models are iterated on 420M images for pre-training stage 1 and 154M images for stage 2. We use the same hyperparameters and number of GPUs for training. The visual encoder and LLM we used for BLIP-2 architecture are Eva-CLIP ViT-g/14~\cite{evaclip} and Vicuna-7b~\citep{vicuna} respectively. More training details are available in Appendix~\ref{appendix:blip2_train} Table~\ref{tab:blip2_setting}.

\textbf{Results.} Two BLIP-2 models pre-trained on different filtered datasets are evaluated on VQAv2~\citep{vqav2} and GQA~\citep{hudson2019gqa} datasets in zero-shot manner and the results of zero-shot VQA performance are shown in Table~\ref{tab:blip2}. The BLIP-2 pre-trained with \ourf{} filtered image-text data achieves +1.7 and + 1.4 improvements on VQAv2 and GQA datasets than the BLIP-2 pre-trained on CLIPSCore filtered dataset.

\begin{table*}[t]
\centering
\begin{tabular}{l l p{10mm}cp{10mm}c}
\hline
\toprule
\textbf{Filter} &\textbf{Metric} & \textbf{VQA} & \textbf{GQA} \\ 
\midrule
CLIPScore  & CLIPScore & 55.1 & 34.8\\
\midrule
\ourf{} & ODF & 56.8 & 36.2 \\
\bottomrule
\hline
\end{tabular}
\caption{
Zero-shot VQA performance of BLIP-2 models pre-trained on dataset filtered by different filtering methods.
}
\label{tab:blip2}
\end{table*}

\subsection{Correlation with Human Scoring}

We follow \cite{g4veval} to compute the correlation between human scoring and model scoring to evaluate the alignment between human and the filtering model. A set of 100 image-text pairs are sampled from CC12M and MSCOCO~\citep{mscoco} and labeled with human scores in terms of the image-text matching. CLIPScore and fine-tuned MLM filters are used to generate the image-text matching scores for the selected image-text pairs. Then, the Pearson and Spearman scores are reported between the human scores and model scores, as presented in Table~\ref{tab:correlation}. Our proposed \textsc{MLM-Filter} scores are significantly aligned and correlated with human quality scores, while CLIPScore does not demonstrate such correlations. The two quality metrics Image-Text Matching and Object Detail Fulfillment all demonstrate significant correlations in similar levels.

\begin{table*}[t]
\centering
\small
\begin{tabular}{l lp{10mm}cp{10mm}c}
\hline
\toprule
\textbf{Filter} &\textbf{Metric} & \textbf{Pearson} & \textbf{Spearman} \\ 
\midrule
CLIPScore  & - & 0.164 & 0.072 \\
\midrule
\ourf{}& ITM & \textbf{0.452}$^*$ & \textbf{0.430}$^*$ \\
\ourf{} & ODF & 0.410$^*$ & 0.384$^*$ \\
\ours{} & ITM & 0.328$^*$ & 0.331$^*$ \\
\ours{} & ODF & 0.368$^*$ & 0.374$^*$ \\
\bottomrule
\hline
\end{tabular}
\caption{
Pearson and Spearman correlations between human-labeled quality scores and scores generated by MLM-Filter and CLIP. Images are scored on a scale of 100 for our MLMFilter, while CLIPScore is also normalized to the scale of 100. The $^{*}$ denotes significant correlations at $p < 0.05$.
}
\label{tab:correlation}
\end{table*}

\subsection{Analysis}

\begin{figure}[ht] 
\centering 
\includegraphics[width=0.5\textwidth]{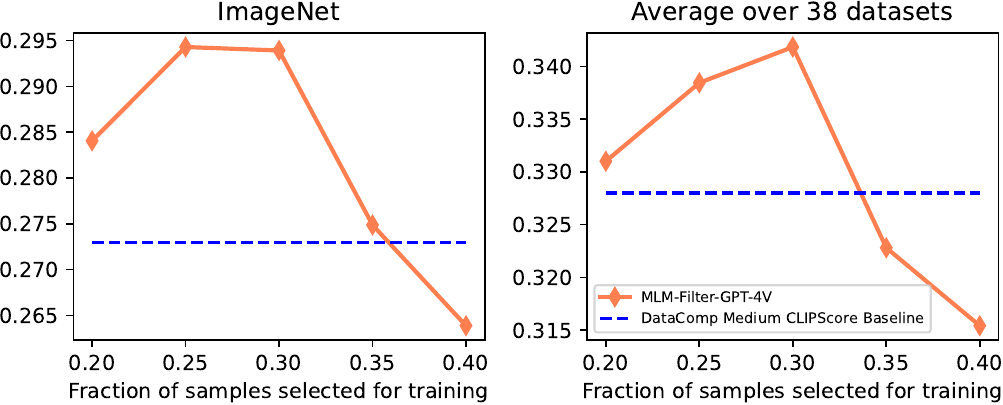}
\caption{Effects of fraction of images selected for training CLIP.}
\label{fig:fraction}
\end{figure}

\textbf{Effects of filtering fraction.} We perform an ablation study to investigate the effects of the fraction of samples selected for pre-training CLIP on DataComp Medium benchmark performance. We select five fractions $\{0.2,0.25,0.3,0.35,0.4\}$ of the total 128M images of DataComp medium pool. The results are presented in Table~\ref{fig:fraction}. The top-30\% of images selected for CLIP training achieve the best performance, which is also observed in \cite{datacomp}. Even adding 5\% poison data leads to a huge performance drop on both ImageNet and average over 38 datasets. 

\textbf{Efficiency of MLM Filters.} 
The MLM Filter used for quality score generation is LLaVA-1.5 with 14B model parameters , while CLIPScore adopts a CLIP ViT-L/14 model with 492M parameter in total. Even if the model size of the proposed MLM Filter is much larger than that of CLIPScore, due to the computation redundancy of the CLIP's dual-encoder architecture, the timecost for generating scores for 10k image-text pairs is average 24.3 mins for MLM Filter versus 11.2 mins for CLIPScore-ViT/L using one A100 GPU. Additionally, with the help of the latest techniques in language model inference acceleration, the TensorRT-LLM toolkit\footnote{https://github.com/NVIDIA/TensorRT-LLM}, we accelerate the score generation of our MLM Filter 4 times over, resulting in 6.1 mins in average for 10k samples. Thus, the proposed MLM Filter can achieve much better efficiency than CLIPScore.


\section{Conclusion}
We propose to instruction-tune Multimodal Language Model on quality scoring tasks and further leverage these fine-tuned MLM as effective data filters to select high-quality image-text pairs from large-scale web-crawled dataset. We find that, on CLIP and BLIP-2 models, pre-training on datasets filtered by our proposed MLM Filter significantly outperforms pre-training on CLIPScore-filtered datasets, demonstrating the superiority of our proposed MLM Filter over CLIPScore filtering.

{\small
\bibliography{example}

\newcommand{\etalchar}[1]{$^{#1}$}
\begin{thebibliography}{GKSS{\etalchar{+}}17}

\bibitem[ADL{\etalchar{+}}22]{flamingo}
Jean-Baptiste Alayrac, Jeff Donahue, Pauline Luc, Antoine Miech, Iain Barr, Yana Hasson, Karel Lenc, Arthur Mensch, Katherine Millican, Malcolm Reynolds, et~al.
\newblock Flamingo: a visual language model for few-shot learning.
\newblock {\em Advances in Neural Information Processing Systems}, 35:23716--23736, 2022.

\bibitem[BGJ{\etalchar{+}}23]{dalle3}
James Betker, Gabriel Goh, Li~Jing, Tim Brooks, Jianfeng Wang, Linjie Li, Long Ouyang, Juntang Zhuang, Joyce Lee, Yufei Guo, et~al.
\newblock Improving image generation with better captions.
\newblock {\em Computer Science. https://cdn. openai. com/papers/dall-e-3. pdf}, 2(3), 2023.

\bibitem[BPK{\etalchar{+}}22]{coyo700m}
Minwoo Byeon, Beomhee Park, Haecheon Kim, Sungjun Lee, Woonhyuk Baek, and Saehoon Kim.
\newblock Coyo-700m: Image-text pair dataset.
\newblock \url{https://github.com/kakaobrain/coyo-dataset}, 2022.

\bibitem[CLD{\etalchar{+}}23]{chen2023sharegpt4v}
Lin Chen, Jisong Li, Xiaoyi Dong, Pan Zhang, Conghui He, Jiaqi Wang, Feng Zhao, and Dahua Lin.
\newblock Sharegpt4v: Improving large multi-modal models with better captions.
\newblock {\em arXiv preprint arXiv:2311.12793}, 2023.

\bibitem[CLL{\etalchar{+}}23]{vicuna}
Wei-Lin Chiang, Zhuohan Li, Zi~Lin, Ying Sheng, Zhanghao Wu, Hao Zhang, Lianmin Zheng, Siyuan Zhuang, Yonghao Zhuang, Joseph~E. Gonzalez, Ion Stoica, and Eric~P. Xing.
\newblock Vicuna: An open-source chatbot impressing gpt-4 with 90\%* chatgpt quality, March 2023.

\bibitem[CLY{\etalchar{+}}23]{chen2023alpagasus}
Lichang Chen, Shiyang Li, Jun Yan, Hai Wang, Kalpa Gunaratna, Vikas Yadav, Zheng Tang, Vijay Srinivasan, Tianyi Zhou, Heng Huang, et~al.
\newblock Alpagasus: Training a better alpaca with fewer data.
\newblock {\em arXiv preprint arXiv:2307.08701}, 2023.

\bibitem[CRLB18]{rationalization}
Oana-Maria Camburu, Tim Rockt{\"a}schel, Thomas Lukasiewicz, and Phil Blunsom.
\newblock e-snli: Natural language inference with natural language explanations.
\newblock {\em Advances in Neural Information Processing Systems}, 31, 2018.

\bibitem[DDS{\etalchar{+}}09]{deng2009imagenet}
Jia Deng, Wei Dong, Richard Socher, Li-Jia Li, Kai Li, and Li~Fei-Fei.
\newblock Imagenet: A large-scale hierarchical image database.
\newblock In {\em 2009 IEEE conference on computer vision and pattern recognition}, pages 248--255. Ieee, 2009.

\bibitem[DLL{\etalchar{+}}23]{instructblip}
Wenliang Dai, Junnan Li, Dongxu Li, Anthony Meng~Huat Tiong, Junqi Zhao, Weisheng Wang, Boyang Li, Pascale Fung, and Steven Hoi.
\newblock Instructblip: Towards general-purpose vision-language models with instruction tuning, 2023.

\bibitem[FJJ{\etalchar{+}}23]{fang2023data}
Alex Fang, Albin~Madappally Jose, Amit Jain, Ludwig Schmidt, Alexander Toshev, and Vaishaal Shankar.
\newblock Data filtering networks.
\newblock {\em arXiv preprint arXiv:2309.17425}, 2023.

\bibitem[GIF{\etalchar{+}}23]{datacomp}
Samir~Yitzhak Gadre, Gabriel Ilharco, Alex Fang, Jonathan Hayase, Georgios Smyrnis, Thao Nguyen, Ryan Marten, Mitchell Wortsman, Dhruba Ghosh, Jieyu Zhang, et~al.
\newblock Datacomp: In search of the next generation of multimodal datasets.
\newblock {\em arXiv preprint arXiv:2304.14108}, 2023.

\bibitem[GKSS{\etalchar{+}}17]{vqav2}
Yash Goyal, Tejas Khot, Douglas Summers-Stay, Dhruv Batra, and Devi Parikh.
\newblock Making the v in vqa matter: Elevating the role of image understanding in visual question answering.
\newblock In {\em Proceedings of the IEEE conference on computer vision and pattern recognition}, pages 6904--6913, 2017.

\bibitem[HDW{\etalchar{+}}24]{kosmos-1}
Shaohan Huang, Li~Dong, Wenhui Wang, Yaru Hao, Saksham Singhal, Shuming Ma, Tengchao Lv, Lei Cui, Owais~Khan Mohammed, Barun Patra, et~al.
\newblock Language is not all you need: Aligning perception with language models.
\newblock {\em Advances in Neural Information Processing Systems}, 36, 2024.

\bibitem[HHF{\etalchar{+}}21]{hessel2021clipscore}
Jack Hessel, Ari Holtzman, Maxwell Forbes, Ronan~Le Bras, and Yejin Choi.
\newblock Clipscore: A reference-free evaluation metric for image captioning.
\newblock {\em arXiv preprint arXiv:2104.08718}, 2021.

\bibitem[HM19]{hudson2019gqa}
Drew~A Hudson and Christopher~D Manning.
\newblock Gqa: A new dataset for real-world visual reasoning and compositional question answering.
\newblock In {\em CVPR}, 2019.

\bibitem[JYX{\etalchar{+}}21]{jia2021scaling}
Chao Jia, Yinfei Yang, Ye~Xia, Yi-Ting Chen, Zarana Parekh, Hieu Pham, Quoc Le, Yun-Hsuan Sung, Zhen Li, and Tom Duerig.
\newblock Scaling up visual and vision-language representation learning with noisy text supervision.
\newblock In {\em International conference on machine learning}, pages 4904--4916. PMLR, 2021.

\bibitem[LLLL23]{llava2}
Haotian Liu, Chunyuan Li, Yuheng Li, and Yong~Jae Lee.
\newblock Improved baselines with visual instruction tuning.
\newblock {\em arXiv preprint arXiv:2310.03744}, 2023.

\bibitem[LLSH23]{blip2}
Junnan Li, Dongxu Li, Silvio Savarese, and Steven Hoi.
\newblock Blip-2: Bootstrapping language-image pre-training with frozen image encoders and large language models.
\newblock {\em arXiv preprint arXiv:2301.12597}, 2023.

\bibitem[LLWL23]{llava}
Haotian Liu, Chunyuan Li, Qingyang Wu, and Yong~Jae Lee.
\newblock Visual instruction tuning.
\newblock {\em arXiv preprint arXiv:2304.08485}, 2023.

\bibitem[LMB{\etalchar{+}}14]{mscoco}
Tsung-Yi Lin, Michael Maire, Serge Belongie, James Hays, Pietro Perona, Deva Ramanan, Piotr Doll{\'a}r, and C~Lawrence Zitnick.
\newblock Microsoft coco: Common objects in context.
\newblock In {\em Computer Vision--ECCV 2014: 13th European Conference, Zurich, Switzerland, September 6-12, 2014, Proceedings, Part V 13}, pages 740--755. Springer, 2014.

\bibitem[MGL{\etalchar{+}}23]{maini2023t}
Pratyush Maini, Sachin Goyal, Zachary~C Lipton, J~Zico Kolter, and Aditi Raghunathan.
\newblock T-mars: Improving visual representations by circumventing text feature learning.
\newblock {\em arXiv preprint arXiv:2307.03132}, 2023.

\bibitem[MKBH21]{mishra2021cross}
Swaroop Mishra, Daniel Khashabi, Chitta Baral, and Hannaneh Hajishirzi.
\newblock Cross-task generalization via natural language crowdsourcing instructions.
\newblock {\em arXiv preprint arXiv:2104.08773}, 2021.

\bibitem[MRFM19]{okvqa}
Kenneth Marino, Mohammad Rastegari, Ali Farhadi, and Roozbeh Mottaghi.
\newblock Ok-vqa: A visual question answering benchmark requiring external knowledge.
\newblock In {\em Conference on Computer Vision and Pattern Recognition (CVPR)}, 2019.

\bibitem[MSSC19]{mishra2019ocrvqa}
Anand Mishra, Shashank Shekhar, Ajeet~Kumar Singh, and Anirban Chakraborty.
\newblock Ocr-vqa: Visual question answering by reading text in images.
\newblock In {\em 2019 international conference on document analysis and recognition (ICDAR)}, pages 947--952. IEEE, 2019.

\bibitem[NIW{\etalchar{+}}22]{nguyen2022quality}
Thao Nguyen, Gabriel Ilharco, Mitchell Wortsman, Sewoong Oh, and Ludwig Schmidt.
\newblock Quality not quantity: On the interaction between dataset design and robustness of clip.
\newblock {\em Advances in Neural Information Processing Systems}, 35:21455--21469, 2022.

\bibitem[Ope23]{gpt4v}
OpenAI.
\newblock Gpt-4v(ision) technical work and authors.
\newblock 2023.

\bibitem[OWJ{\etalchar{+}}22]{instructgpt}
Long Ouyang, Jeffrey Wu, Xu~Jiang, Diogo Almeida, Carroll Wainwright, Pamela Mishkin, Chong Zhang, Sandhini Agarwal, Katarina Slama, Alex Ray, et~al.
\newblock Training language models to follow instructions with human feedback.
\newblock {\em Advances in Neural Information Processing Systems}, 35:27730--27744, 2022.

\bibitem[RG20]{sbert}
Nils Reimers and Iryna Gurevych.
\newblock Making monolingual sentence embeddings multilingual using knowledge distillation.
\newblock In {\em Proceedings of the 2020 Conference on Empirical Methods in Natural Language Processing}. Association for Computational Linguistics, 11 2020.

\bibitem[RKH{\etalchar{+}}21]{clip}
Alec Radford, Jong~Wook Kim, Chris Hallacy, Aditya Ramesh, Gabriel Goh, Sandhini Agarwal, Girish Sastry, Amanda Askell, Pamela Mishkin, Jack Clark, Gretchen Krueger, and Ilya Sutskever.
\newblock Learning transferable visual models from natural language supervision.
\newblock In {\em ICML}, 2021.

\bibitem[SBV{\etalchar{+}}22]{laion5b}
Christoph Schuhmann, Romain Beaumont, Richard Vencu, Cade Gordon, Ross Wightman, Mehdi Cherti, Theo Coombes, Aarush Katta, Clayton Mullis, Mitchell Wortsman, et~al.
\newblock Laion-5b: An open large-scale dataset for training next generation image-text models.
\newblock {\em Advances in Neural Information Processing Systems}, 35:25278--25294, 2022.

\bibitem[SDGS18a]{cc3m}
Piyush Sharma, Nan Ding, Sebastian Goodman, and Radu Soricut.
\newblock Conceptual captions: A cleaned, hypernymed, image alt-text dataset for automatic image captioning.
\newblock In {\em Proceedings of the 56th Annual Meeting of the Association for Computational Linguistics (Volume 1: Long Papers)}, pages 2556--2565, 2018.

\bibitem[SDGS18b]{cc12m}
Piyush Sharma, Nan Ding, Sebastian Goodman, and Radu Soricut.
\newblock Conceptual captions: A cleaned, hypernymed, image alt-text dataset for automatic image captioning.
\newblock In {\em Proceedings of the 56th Annual Meeting of the Association for Computational Linguistics (Volume 1: Long Papers)}, pages 2556--2565, 2018.

\bibitem[SFW{\etalchar{+}}23]{evaclip}
Quan Sun, Yuxin Fang, Ledell Wu, Xinlong Wang, and Yue Cao.
\newblock Eva-clip: Improved training techniques for clip at scale.
\newblock {\em arXiv preprint arXiv:2303.15389}, 2023.

\bibitem[SGM{\etalchar{+}}22]{saikh2022scienceqa}
Tanik Saikh, Tirthankar Ghosal, Amish Mittal, Asif Ekbal, and Pushpak Bhattacharyya.
\newblock Scienceqa: A novel resource for question answering on scholarly articles.
\newblock {\em International Journal on Digital Libraries}, 23(3):289--301, 2022.

\bibitem[Sha23]{sharegpt}
Share{GPT}.
\newblock \url{https://sharegpt.com/}, 2023.

\bibitem[SHRS20]{sidorov2020textcaps}
Oleksii Sidorov, Ronghang Hu, Marcus Rohrbach, and Amanpreet Singh.
\newblock Textcaps: a dataset for image captioning with reading comprehension.
\newblock In {\em Computer Vision--ECCV 2020: 16th European Conference, Glasgow, UK, August 23--28, 2020, Proceedings, Part II 16}, pages 742--758. Springer, 2020.

\bibitem[STQ{\etalchar{+}}20]{song2020mpnet}
Kaitao Song, Xu~Tan, Tao Qin, Jianfeng Lu, and Tie-Yan Liu.
\newblock Mpnet: Masked and permuted pre-training for language understanding.
\newblock {\em Advances in Neural Information Processing Systems}, 33:16857--16867, 2020.

\bibitem[SVB{\etalchar{+}}21]{laion400m}
Christoph Schuhmann, Richard Vencu, Romain Beaumont, Robert Kaczmarczyk, Clayton Mullis, Aarush Katta, Theo Coombes, Jenia Jitsev, and Aran Komatsuzaki.
\newblock Laion-400m: Open dataset of clip-filtered 400 million image-text pairs.
\newblock {\em arXiv preprint arXiv:2111.02114}, 2021.

\bibitem[TGZ{\etalchar{+}}23]{alpaca}
Rohan Taori, Ishaan Gulrajani, Tianyi Zhang, Yann Dubois, Xuechen Li, Carlos Guestrin, Percy Liang, and Tatsunori~B. Hashimoto.
\newblock Stanford alpaca: An instruction-following llama model.
\newblock \url{https://github.com/tatsu-lab/stanford_alpaca}, 2023.

\bibitem[TJS23]{tong2023mass}
Shengbang Tong, Erik Jones, and Jacob Steinhardt.
\newblock Mass-producing failures of multimodal systems with language models.
\newblock {\em arXiv preprint arXiv:2306.12105}, 2023.

\bibitem[TLZ{\etalchar{+}}24]{tong2024eyes}
Shengbang Tong, Zhuang Liu, Yuexiang Zhai, Yi~Ma, Yann LeCun, and Saining Xie.
\newblock Eyes wide shut? exploring the visual shortcomings of multimodal llms.
\newblock {\em arXiv preprint arXiv:2401.06209}, 2024.

\bibitem[TMS{\etalchar{+}}23]{llama2}
Hugo Touvron, Louis Martin, Kevin Stone, Peter Albert, Amjad Almahairi, Yasmine Babaei, Nikolay Bashlykov, Soumya Batra, Prajjwal Bhargava, Shruti Bhosale, et~al.
\newblock Llama 2: Open foundation and fine-tuned chat models.
\newblock {\em arXiv preprint arXiv:2307.09288}, 2023.

\bibitem[WBZ{\etalchar{+}}21]{flan}
Jason Wei, Maarten Bosma, Vincent~Y Zhao, Kelvin Guu, Adams~Wei Yu, Brian Lester, Nan Du, Andrew~M Dai, and Quoc~V Le.
\newblock Finetuned language models are zero-shot learners.
\newblock {\em arXiv preprint arXiv:2109.01652}, 2021.

\bibitem[WDC{\etalchar{+}}22]{valm}
Weizhi Wang, Li~Dong, Hao Cheng, Haoyu Song, Xiaodong Liu, Xifeng Yan, Jianfeng Gao, and Furu Wei.
\newblock Visually-augmented language modeling.
\newblock {\em arXiv preprint arXiv:2205.10178}, 2022.

\bibitem[WJHS23]{wei2023instructiongpt}
Lai Wei, Zihao Jiang, Weiran Huang, and Lichao Sun.
\newblock Instructiongpt-4: A 200-instruction paradigm for fine-tuning minigpt-4.
\newblock {\em arXiv preprint arXiv:2308.12067}, 2023.

\bibitem[WLY{\etalchar{+}}23]{cogvlm}
Weihan Wang, Qingsong Lv, Wenmeng Yu, Wenyi Hong, Ji~Qi, Yan Wang, Junhui Ji, Zhuoyi Yang, Lei Zhao, Xixuan Song, Jiazheng Xu, Bin Xu, Juanzi Li, Yuxiao Dong, Ming Ding, and Jie Tang.
\newblock Cogvlm: Visual expert for pretrained language models, 2023.

\bibitem[WWS{\etalchar{+}}22]{cot}
Jason Wei, Xuezhi Wang, Dale Schuurmans, Maarten Bosma, Ed~Chi, Quoc Le, and Denny Zhou.
\newblock Chain of thought prompting elicits reasoning in large language models.
\newblock {\em arXiv preprint arXiv:2201.11903}, 2022.

\bibitem[XXT{\etalchar{+}}23]{xu2023demystifying}
Hu~Xu, Saining Xie, Xiaoqing~Ellen Tan, Po-Yao Huang, Russell Howes, Vasu Sharma, Shang-Wen Li, Gargi Ghosh, Luke Zettlemoyer, and Christoph Feichtenhofer.
\newblock Demystifying clip data.
\newblock {\em arXiv preprint arXiv:2309.16671}, 2023.

\bibitem[YLL{\etalchar{+}}23]{msgpt4v}
Zhengyuan Yang, Linjie Li, Kevin Lin, Jianfeng Wang, Chung-Ching Lin, Zicheng Liu, and Lijuan Wang.
\newblock The dawn of lmms: Preliminary explorations with gpt-4v (ision).
\newblock {\em arXiv preprint arXiv:2309.17421}, 9(1):1, 2023.

\bibitem[YTK{\etalchar{+}}23]{yu2023devil}
Haichao Yu, Yu~Tian, Sateesh Kumar, Linjie Yang, and Heng Wang.
\newblock The devil is in the details: A deep dive into the rabbit hole of data filtering.
\newblock {\em arXiv preprint arXiv:2309.15954}, 2023.

\bibitem[ZCS{\etalchar{+}}23]{zhu2023minigpt}
Deyao Zhu, Jun Chen, Xiaoqian Shen, Xiang Li, and Mohamed Elhoseiny.
\newblock Minigpt-4: Enhancing vision-language understanding with advanced large language models.
\newblock {\em arXiv preprint arXiv:2304.10592}, 2023.

\bibitem[ZLW{\etalchar{+}}23]{g4veval}
Xinlu Zhang, Yujie Lu, Weizhi Wang, An~Yan, Jun Yan, Lianke Qin, Heng Wang, Xifeng Yan, William~Yang Wang, and Linda~Ruth Petzold.
\newblock Gpt-4v (ision) as a generalist evaluator for vision-language tasks.
\newblock {\em arXiv preprint arXiv:2311.01361}, 2023.

\bibitem[ZPK{\etalchar{+}}19]{vtab}
Xiaohua Zhai, Joan Puigcerver, Alexander Kolesnikov, Pierre Ruyssen, Carlos Riquelme, Mario Lucic, Josip Djolonga, Andre~Susano Pinto, Maxim Neumann, Alexey Dosovitskiy, et~al.
\newblock The visual task adaptation benchmark.
\newblock 2019.

\end{thebibliography}
}

\newpage
\appendix
\onecolumn

\section{Prompt Construction}
\label{appendix:prompts}

After manually writing the first version of prompts, we leverage the GPT-4 to refine the human-written prompts. The final prompts for four quality scoring tasks are shown below:
\begin{tcolorbox}[enhanced]
\textbf{Image Text Matching}\vspace{3pt}\\ 
Please evaluate if the provided text caption accurately represents the main features and objects of the image. The caption doesn't need to detail every aspect of the image, but it should capture its primary theme. Rate the overall quality of the text caption's match to the image on a scale of 1-100, considering the criteria mentioned.
\end{tcolorbox}

\begin{tcolorbox}[enhanced]
\textbf{Object Detail Fulfillment}\vspace{3pt}\\ 
Please evaluate the text caption to determine if it provides detailed descriptions of objects that align with the image description. Specifically, assess if the caption sufficiently describes the color, size, position, shape, material, etc., of the objects. Afterward, rate the caption's overall accuracy in capturing object details from the image on a scale of 1-100, based on the criteria provided.
\end{tcolorbox}

\begin{tcolorbox}[enhanced]
\textbf{Caption Text Quality}\vspace{3pt}\\ 
Please evaluate the text caption based on the following criteria: Grammatical Correctness, Diversity of Vocabulary (e.g., the range and uniqueness of words used), Fluency (e.g., smoothness and natural flow of sentences), Readability, Length, and Structure. Assign an overall quality score on a scale of 1-100.
\end{tcolorbox}

\begin{tcolorbox}[enhanced]
\textbf{Semantic Understanding}\vspace{3pt}\\ 
Please evaluate the given text caption in relation to its corresponding image description. Your goal is to determine if the text caption provides additional semantic information that isn't readily apparent just from the image itself.

For example:

1. If the image description mentions "a man" but the caption elaborates he is a "homeless man" or a "businessman," then the caption is enriching the semantic context.

2. If the caption introduces concepts like the mathematical tangent function, which require in-depth knowledge to deduce, it is imparting external semantics.

3. Captions revealing specific location addresses, festival details, or other nuanced data not easy to infer from the image also provide external semantic information.

4. Directly identifying specific entities in the image such as buildings, people, bird species, animal breeds, car models, engines, etc., in the caption introduces additional insights.

5. Should the image act as a contextual backdrop and the caption describes elements not explicitly showcased in the image, it has semantic depth.

6. Lastly, if the caption depicts relationships between the subjects in the image, which need commonsense knowledge to understand, it should be considered semantically rich.

Please assess and determine the extent of semantic enrichment the caption provides over the image description. Rate the text caption's semantic depth on a scale from 1 to 100.
\end{tcolorbox}


\newpage

\section{Examples for Two Prompting Strategies}
\label{appendix:prompt_strategy}
\begin{table*}[h]
\small
\centering
\begin{tabular}{p{0.45\textwidth}|p{0.45\textwidth}}
\toprule
Example for \bred{Chain-of-Thought Reasoning} &
Example for \bred{Rationalization}
\\
\midrule
Please evaluate if the provided text caption accurately represents the main features and objects of the image. The caption doesn't need to detail every aspect of the image, but it should capture its primary theme. Rate the overall quality of the text caption's match to the image on a scale of 1-100, considering the criteria mentioned.
&
Please evaluate if the provided text caption accurately represents the main features and objects of the image. The caption doesn't need to detail every aspect of the image, but it should capture its primary theme. Rate the overall quality of the text caption's match to the image on a scale of 1-100, considering the criteria mentioned.
\\
\vspace{3pt}\bblack{Please} \bblue{think step by step to first output your reasons} \bblack{to give such a score. In the subsequent line, please output a single line containing the value indicating the scores.}
&
\vspace{3pt}\bblack{Please} \bblue{first output a single line containing the value indicating the scores}. \bblack{In the subsequent line, please} \bblue{provide a comprehensive explanation of your evaluation}, \bblack{avoiding any potential bias.}
\\
\bottomrule
\hline
\end{tabular}
\caption{Prompts for zero-shot Chain-of-Thought reasoning and Rationalization reasoning for assessing the image-text matching score.}
\label{tab:prompt_strategy}
\end{table*}

\section{Sampling Final Instructions for Scoring Tasks}
\label{appendix:sampling}

\begin{lstlisting}[language=Python]
final_data = []
buckets = 10 * [[]]
for d in data:
    buckets[d['score']//10].append(d)
threshold = 130
num_downsample_bucket = 0
for b in buckets:
    if len(b) < threshold:
        final_data += p
    else:
        num_downsample_bucket += 1
downsampling_num = 1000 - len(final_data)
num_per_bucket = downsampling_num // num_downsample_bucket
for b in buckets:
    if len(b) > threshold:
        final_data += random.sample(b, num_per_bucket)
total_data += final_data[:1000]
\end{lstlisting}

\section{Data Mixture of Multi-Task Multimodal Instructions}
\label{appendix:data_mixture}

\begin{table*}[h!]
\centering
\small
\scalebox{1}{
\begin{tabular}{l| ll}
\hline
\toprule
Data & Size & Task \\
\midrule
Visual Conversation~\citep{llava} & 5K & Conversation \\
Complex Reasoning~\citep{llava}  & 16k & Visual Reasoning \\
Detail Description~\citep{llava}  & 5k & Captioning \\
\midrule
ShareGPT~\citep{sharegpt} & 10K & Language-Only Instructions \\
\midrule
VQAv2~\citep{vqav2} & 2K & VQA \\
GQA~\citep{hudson2019gqa} & 3K & Visual Reasoning \\
OKVQA~\citep{okvqa} & 2K & Knowledge Grounded VQA \\
OCRVQA~\citep{mishra2019ocrvqa} & 1K & OCR \\
TextCaption~\citep{sidorov2020textcaps} & 2K & Captioning \\
\midrule
ITM Scoring & 1k & \multirow{4}{*}{Data Quality Scoring}  \\
ODF Scoring & 1k &  \\
CTQ Scoring & 1k & \\
SU Scoring & 1k &  \\
\midrule
Total & 50k & \\
\bottomrule
\hline
\end{tabular}
}
\caption{
Multimodal instruction data mixture of the data quality scoring tasks and other multimodal tasks.
}
\label{tab:data_mixture}
\end{table*}

\section{BLIP-2 Training Details}
\label{appendix:blip2_train}

The detailed training hypperparamters and settings of BLIP-2 stage 1 and stage 2 are presented in Table~\ref{tab:blip2_setting}.

\begin{table*}[ht]
\begin{center}
\begin{tabular}{l|c}
\hline
\toprule
\textbf{Hyperparameter}  & BLIP-2 \\ 
\midrule
\multicolumn{2}{c}{\textbf{Stage-1 Pre-training}} \\
\# Trainable Parameters & 188M \\  
Precision & \texttt{float16}  \\ 
Global Batch Size & 1680 \\ 
\# Training Steps & 250k \\
\# GPUs & 16 \\
\# Gradient Accumulation Steps & 1 \\
Min LR & 1e-5 \\
Peak LR & 1e-4 \\
\# Warmup Steps & 2000 \\
LR Scheduler & Cosine LR Decay \\
Weight Decay & 0.05 \\
Adam $(\beta_1, \beta_2)$ & (0.9, 0.98) \\   
\midrule
\multicolumn{2}{c}{\textbf{Stage-2 Pre-training}} \\
\# Trainable Parameters & 188M \\  
Precision & \texttt{float16}  \\ 
Global Batch Size & 1920 \\
\# Training Steps & 80k \\
\# GPUs & 16 \\
\# Gradient Accumulation Steps & 4 \\
Min LR & 5e-5 \\
Peak LR & 1e-4 \\
\# Warmup Steps & 2000 \\
LR Scheduler & Cosine LR Decay \\
Weight Decay & 0.05 \\
Adam $(\beta_1, \beta_2)$ & (0.9, 0.98) \\   
\bottomrule
\hline
\end{tabular}
\caption{Training details for BLIP-2 pre-training stage 1 and stage 2.}
\label{tab:blip2_setting}
\end{center}
\end{table*}

\end{document}